\documentclass[11pt,a4paper]{article}
\usepackage[hyperref]{acl2019}
\usepackage{times}
\usepackage{latexsym}
\usepackage{graphicx}
\usepackage{graphics}
\usepackage{subfigure}
\usepackage{amsfonts}
\usepackage{url}
\usepackage{multirow}
\usepackage{graphicx}
\usepackage{booktabs}

\usepackage{url}
\usepackage{epsfig}
\usepackage{enumitem}
\setenumerate[1]{itemsep=0pt,partopsep=0pt,parsep=\parskip,topsep=5pt}
\setitemize[1]{itemsep=0pt,partopsep=0pt,parsep=\parskip,topsep=5pt}
\setdescription{itemsep=0pt,partopsep=0pt,parsep=\parskip,topsep=5pt}
\usepackage[font=small,labelfont=bf,textfont=md]{caption}

\aclfinalcopy % Uncomment this line for the final submission
\title{KCAT: A Knowledge-Constraint Typing Annotation Tool}
\author{Sheng Lin$^1$, Luye Zheng$^1$, Bo Chen$^1$, Siliang Tang$^1$\thanks{Corresponding Author.}, Yueting Zhuang$^1$, \\
    \textbf{Fei Wu$^1$, Zhigang Chen$^2$, Guoping Hu$^2$ \& Xiang Ren$^3$}\\
    $^1$Zhejiang University\\ 
    $^2$iFLYTEK Research, $^3$University of Southern California, \\
    \texttt{\{shenglin, antlar, chenbo123\}@zju.edu.cn},\\
    \texttt{\{siliang, yzhuang, wufei\}@zju.edu.cn},\\
    \texttt{\{zgchen, gphu\}@iflytek.com},\\ \texttt{xiangren@usc.edu}}
\date{}

\begin{document}
\maketitle
\begin{abstract}
Fine-grained Entity Typing is a tough task which suffers from noise samples extracted from distant supervision. Thousands of manually annotated samples can achieve greater performance than millions of samples generated by the previous distant supervision method. Whereas, it's hard for human beings to differentiate and memorize thousands of types, thus making large-scale human labeling hardly possible. In this paper, we introduce a Knowledge-Constraint Typing Annotation Tool (KCAT\footnote{Code is available at https://github.com/donnyslin/KCAT}), which is efficient for fine-grained entity typing annotation. KCAT reduces the size of candidate types to an acceptable range for human beings through entity linking and provides a Multi-step Typing scheme to revise the entity linking result. Moreover, KCAT provides an efficient Annotator Client to accelerate the annotation process and a comprehensive Manager Module to analyse crowdsourcing annotations. Experiment shows that KCAT can significantly improve annotation efficiency, the time consumption increases slowly as the size of type set expands. %Besides, KCAT ensures the high quality of annotation results.
% In this paper, we introduce a Fine-grained Entity Typing Annotation Tool, KCAT, which is efficient and comprehensive. KCAT provides many shortcut keys and modification operation for annotator convenience. Through entity linking, KCAT filters out the irrelevant types and makes it more easy to find the target type. The incorrect entity linking result can be revised by the interaction of the two tasks. Experiments have shown that our toolkit can improve the efficiency compared to the traditional method, while ensuring high quality annotation results.To the best of our knowledge,we are the first one to implement fine-grained entity typing annotation tool.
\end{abstract}

\section{Introduction}
    Recent years Natural Language Processing community has seen a  surge  of interests  in fine-grained  entity typing (\textbf{FET}) as it serves as an important cornerstone of  several nature language processing tasks including relation extraction ~\cite{mintz2009distant}, entity linking ~\cite{raiman2018deeptype}, and knowledge base completion \cite{dong2014knowledge}. Given an entity mention (i.e. a sequence of token spans representing an entity) in the corpus, \textbf{FET} aims at uncovering its context-dependent type. Table \ref{type being reduced} includes Fine-grained Entity Typing datasets in recent years, the target types often form a type hierarchy.

% \begin{figure}[h]
% \centering
% \includegraphics[width=0.48\textwidth, angle=0]{figures/Linking_ver1.pdf}
% \caption{Framework of KCAT}
% \label{linking of KCAT}
% \end{figure}

% \begin{figure*}[t]
% \centering
% \includegraphics[width=1.0\textwidth, angle=0]{figures/Annotator_ver1.pdf}
% % \includegraphics[width=0.50\textwidth, angle=0]{figures/framework_luye_ver4.pdf}
% \caption{Framework of KCAT}
% \label{framework of KCAT}
% \end{figure*}
    The difficulty of FET and FET Annotation both increase rapidly with the growth of type hierarchy's depth. Previous research work mainly focus on generating train corpus with distant supervision \cite{ling2012fine,gillick2014context,ren2016afet,choi2018ultra}. In spite of its efficiency, distant supervision brings the problem of noisy labels, for example,
    %as shown in Figure \ref{EL}, 
    \{\textit{Other, brand}\} are noisy labels for \textit{`Kobe'} in \textit{``\underline{Kobe} scored 60 points in the final game.''}. According to \cite{choi2018ultra}, 6000 manually labeled samples achieved greater performance than millions of samples generated by distant supervision. \cite{Onoe_Durrett_2019} observed that noisy samples may even cause damage to the performance of FET model. \cite{ren2016label,Onoe_Durrett_2019} proposed  label noise reduction methods, which are pretty complicated and hard to migrate.
    
    % \footnote{Code is avaiable at URL}
    % \vspace*{-5pt}
    % \begin{table}[]
    % \small
    % \centering
    % \begin{tabular}{@{}lrr@{}}
    % \toprule
    % \textbf{Dataset}   & \textbf{\#types} & \textbf{\#depths} \\ \midrule
    % BBN\cite{ren2016afet} & 47      & 2        \\
    % OntoNotes\cite{gillick2014context} & 89      & 3        \\
    % FIGER\cite{ling2012fine}     & 112     & 2        \\
    % %HYENA\cite{yosef2012hyena}     & 505     & 9        \\
    % %Freebase\cite{bollacker2008freebase}  & 2k      & 2        \\
    % %TypeNet\cite{murtyfiner}   & 1941    & 14       \\
    %     YAGO\cite{nlp.cs.rpi.edu}  & 7309      & 14        \\
    % WordNet\cite{fellbaum2012wordnet}   & 16k     & 14       \\
    % \bottomrule
    % \end{tabular}
    % \caption{\label{font-table} Statistics of Fine-Grained Typing Datasets}
    % \label{type dataset}
    % \end{table}
    Thus the annotation corpus for FET is important and necessary. However, it is not easy to annotate a corpus for FET since it's hard for human beings to differentiate and memorize thousands of types.
    
    %Draw on the experience of distant supervision, 
    To solve this extremely hard annotation task, we use Entity Linking~(EL) to constrain the candidate types of the entity mention. Entity Linking, which tries to link entity mention to a unique entity in a specific knowledge base (i.e. Yago or Freebase), has been studied for years. The state-of-the-art EL system yields 0.93 F1 score in Conll2003\cite{sang2000introduction}, while the F1 scores of FET vary from 0.40 \cite{Onoe_Durrett_2019} to 0.79 \cite{abhishek-anand-awekar:2017:EACLlong} on different datasets. With the help of EL, the candidate types of a mention can be greatly reduced as shown in the Table \ref{type being reduced}. Based on this observation, we develop a Knowledge-Constraint Typing Annotation Tool~(KCAT). KCAT uses an external entity linking tool to constrain the candidate types of mentions. Because errors made by Entity Linking are inevitable, we provide an EL revision extension in KCAT which can also help the annotation of Entity Linking. The extension uses the coarse-grained type of mention to constrain the candidate entities of mention, which greatly saves the time of revising EL result. The details of the annotation interaction are described in section \ref{annotation interaction}.
    Besides using the Knowledge Constraint technology to make the annotation of FET easier, KCAT provides other 4 functions to further improve the annotation efficiency: Hierarchical Structure Visualization, Annotation Hint, Annotation Modification and Annotation Export. In brief, KCAT has the advantages as follows:
\begin{itemize}[leftmargin=*]
% \setlength{\itemsep}{0pt}
% \setlength{\parsep}{0pt}
% \setlength{\parskip}{0pt}
    % \item\textbf{Intelligent Recommendation:} it provide type candidates that are reduced after linking and visualize the type structure; the entity description of the wiki and types description are also showed on the interface; support multiple shortcuts.
    \item\textbf{Knowledge-Constraint: }it visualizes candidate types hierarchically, which is extracted from Knowledge Base and reduced by entity linking.
    \item\textbf{Efficient:} it supports multiple shortcuts to improve annotation efficiency; entity description of the wiki and type description to help distinguish candidate types.
    \item\textbf{Portable:} it is only required to replace a few json files to complete the migration of different datasets.
    % \item\textbf{Portable:} it doesn't require many configurations, only need python3 and tkinter libraries and does not rely on specific operate system.
    \item\textbf{Comprehensive: }it supports crowdsourcing results comparison and integration. 
\end{itemize}

The rest of the paper is organized as: Section \ref{related work} briefly describes recent research in FET. Section \ref{annotation interaction} introduces the overview of our framework. Section \ref{KCAT ARC} describes the architecture of KCAT and its detail functions. Section \ref{experiment} analyses the efficiency comparison and annotation quality in different annotation mode.  Finally, Section \ref{conclusion} concludes this paper.

\begin{table}[]
\small
\centering
\begin{tabular}{@{}lrrrr@{}}
\toprule
\textbf{Dataset}   & \textbf{Depth}  & \textbf{\#Types} & \textbf{\#KC Types} & \textbf{Ratio.\%} \\ \midrule
BBN & 2  & 47   & 2.1 &       4.3\%\\
FIGER    & 2  & 110   & 2.3 &   2.1\% \\
TAC 2018   & 14  &7309  & 8.5 & 0.1\%    \\  \bottomrule
\end{tabular}
%修改一下这个caption使得这些数字的意思更加清晰
\caption{Size of Candidate Types before and after \textbf{Knowledge-Constraint} on Different Datasets, and the ratio of the latter to the former}
\end{table}\label{type being reduced}

% \begin{figure}[h]
% \centering
% \includegraphics[width=0.50\textwidth, angle=0]{figures/interaction_new.pdf}
% \caption{interaction between typing and linking}
% \label{interaction}
% \end{figure}

% \begin{figure*}[t]
% \centering
% \includegraphics[width=0.95\textwidth, angle=0]{figures/screenshot_v1.pdf}
% \caption{Screenshot of KCAT (different parts are seperated by bold red line)}
% \label{screenShot}
% \end{figure*}

% \begin{figure}[h]
% \centering
% \includegraphics[width=0.50\textwidth, angle=0]{figures/interaction.pdf}
% \caption{Interaction between Typing and Linking}
% \label{interaction}
% \end{figure}
\section{Related Works}\label{related work}
% \begin{itemize}
    % \item
    Named Entity Recognition~(NER) has been studied for several decades,  which classifies  coarse-grained  types (e.g.   person,  location). In order to reduce the cost of obtaining fine-grained typing corpus, distant supervision has been widely used in FET \cite{ling2012fine,gillick2014context,ren2016afet,choi2018ultra}. Inevitably, distant supervision brings the unique challenge of noisy labels in FET which seriously slows down the research process in this field. Many researchers focus on noise reduction of label\cite{ren2016label,ren2016afet,Onoe_Durrett_2019}. The costliness of annotated corpus and the problem of noisy labels greatly hurt the usability of FET technique. Previous entity typing annotation tools~\cite{stenetorp2012brat,yang2017yedda} focus on the coarse-grained types and is hard to migrate to fine-grained types. 
    %Recently,\cite{yang2017yedda} released a local NER annotation tool which only support up to ten types. 
     A specific and carefully designed annotation tool is urgently needed for FET.
    %  , which finally contribute to the birth of KCAT.
    To the best of our knowledge, KCAT is the first Fine-grained Entity Typing Annotation tool. \\%Figure~\ref{framework of KCAT} shows the details about our framework.
    % \item 
    %\textbf{Entity Linking: }Entity Linking (EL) is critical to build connection between text corpus and knowledge base. Traditional Entity Linking models are based on local decision method by extracted some local features~\cite{bunescu2006using, cucerzan2007large, raiman2018deeptype}. However, these methods ignore the entity-to-enity coherence, hence global methods attract considerable  attention in the past few  years~\cite{ganea2017deep,le2018improving,yang2018collective}, which yield better performance than local methods. Recently, reinforcement learning is applied into entity linking ~\cite{fang2019joint}, which regards entity linking as a sequence decision problem and achieves state-of-the-art result.
% \end{itemize}
\begin{figure}[h]
\centering
\includegraphics[width=0.48\textwidth, angle=0]{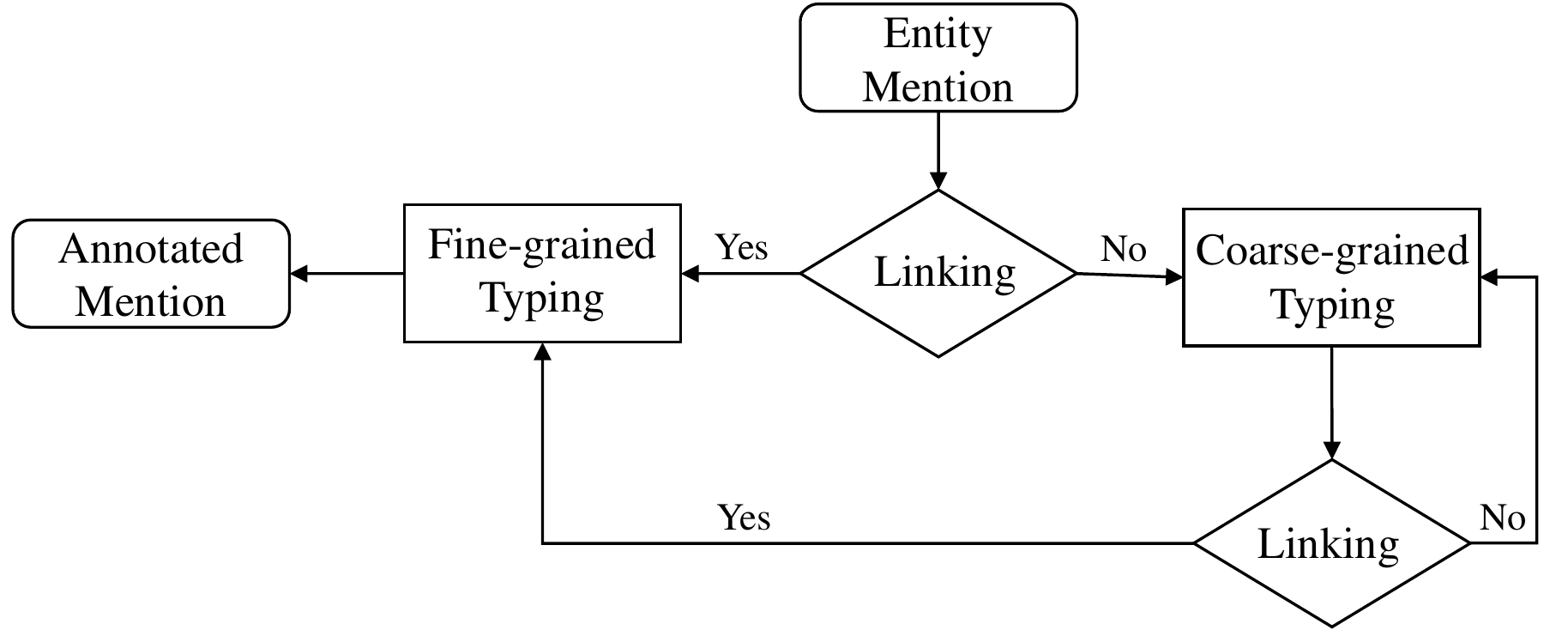}
\caption{The framework of KCAT}
\label{framework of KCAT}
\end{figure}
\section{Overview}\label{annotation interaction}
This section overviews the proposed framework as shown in Figure \ref{framework of KCAT}. KCAT leverages \textbf{Type Hierarchy from Knowledge Base}, and reduces the size of candidate types to a small range through \textbf{Entity Linking}. Furthermore, KCAT proposes a \textbf{Multi-step Typing} scheme because the result from Entity Linking may be incorrect. As shown in Figure \ref{framework of KCAT}, given an entity mention, KCTA links it to entity in Knowledge Base by EL model. Fine-grained type can be directly labeled if this result from model is correct, otherwise KCTA provides entity linking revision by coarse-grained type constraint to filter out candidates entities with inconsistent types and finally labels fine-grained type. The details will be described following.

\subsection{Type Hierarchy and Knowledge Base} Given a set of types $\mathcal{T} = \{t_1, ..., t_N\}$, these types usually form a  Directed Acyclic Graph~(DAG) or more commonly a tree.
%, since child type may have one or more parent type, among which tree structure is the commonest. 
Each entity $e$ in Knowledge Base $\mathcal{K}$ has only several types, $\mathcal{T}_e= \{t_1, ..., t_M\}\subset \mathcal{T}$. In general, the size $|\mathcal{T}_e|$ is far less than the size $|\mathcal{T}|$. As shown in Table \ref{type being reduced}, the average size $\overline{|\mathcal{T}_e|}$ % in $\mathcal{K}$
maintains in a small range as $|\mathcal{T}|$ expands. %, which is only related with the hierarchical depth. 
Therefore, EL, as a Knowledge Constraint method% linking an enity mention to entity in $\mathcal{K}$, called Entity Linking, 
, helps to reduce the candidate size significantly.
\subsection{Entity Linking}
\begin{figure}[h]
\centering
\includegraphics[width=0.46\textwidth, angle=0]{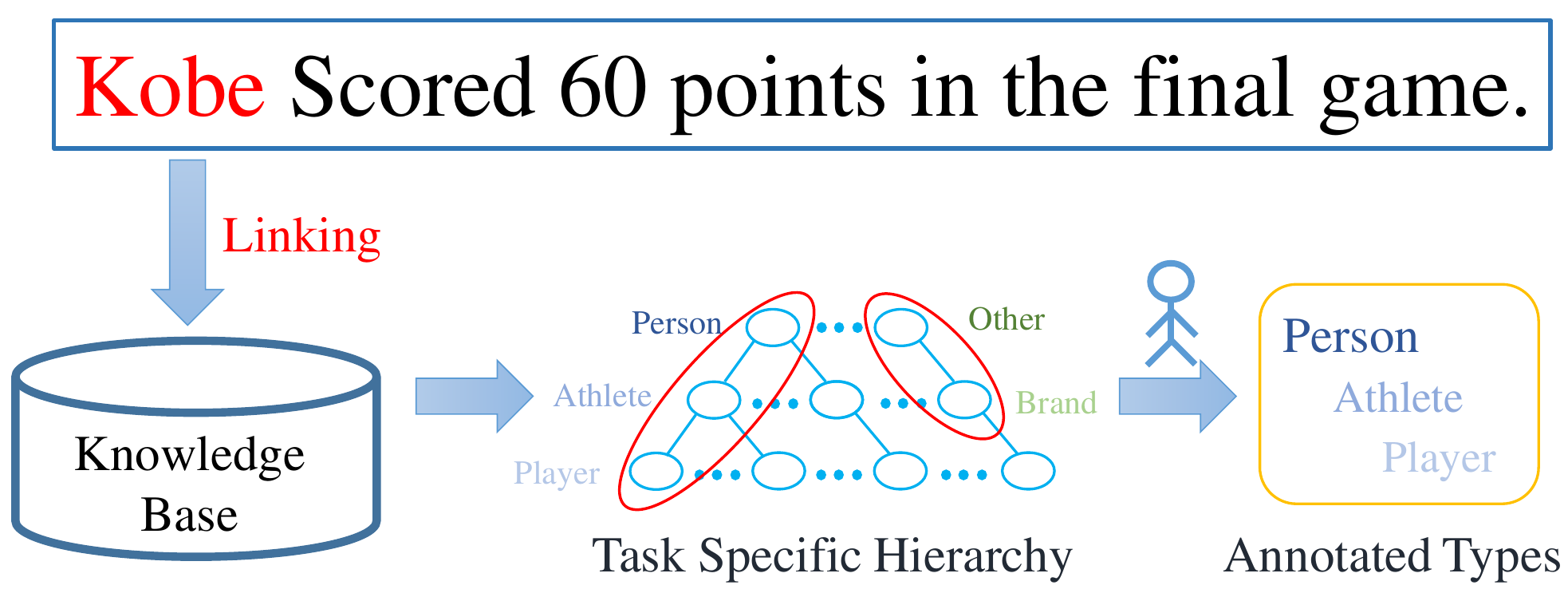}
\caption{The Process of Knowledge Constraint by Entity Linking}
\label{EL}
\end{figure}

Given a set of entity mentions $\mathcal{M} = \{m_1, ..., m_T\}$ in corpus $\mathcal{D}$, Entity Linking aims to link each mention $m_t$ to its corresponding gold entity $e_i^*$ in $\mathcal{K}$. Such process is usually divided into two steps:
\emph{Candidate generation} first collects a set of possible candidate entities $\mathcal{E}_i = \{e_i^1, ..., e_i^{|\mathcal{E}_i|}\}$ for $m_i$;
\emph{Candidate ranking} is then applied to rank all candidates. The linking system selects the top ranked candidate as the predicted entity $\hat{e}_i$.
Given a mention in text, our system firstly links it to $\mathcal{K}$ by state-of-the-art Entity Linking system~\cite{le2018improving}, which yields 0.93 F1 score on Conll 2003 dataset. 
% This process provides Knowledge Constraint for FET, 
As shown in Figure \ref{EL}, \emph{Kobe} is an entity mention which can be linked to ``Kobe Bean Bryant'' in $\mathcal{K}$, whose types only contain \emph{person} and its descendant, we only need to type on the subtree. 
Whereas, after EL there are still 7\% wrong linked mentions which need to be manually revised. Hence, for each mention, we provide the candidate entities with top 20 ranked score from EL system as the revision choices. In current EL systems, ground truth entity coverage can reach 0.98~\cite{ganea2017deep} which ensuring the recall of revision choices. 

Even though, the revision progress can be tough as The context may be ambiguous for manually linking mentions. Distinguishing different entities can be time-consuming and difficult for an amateurish annotator. To accelerate the revision progress, KCAT uses multi-step typing to reduce the number of candidate entities. %Due the coarse-grained type annotation is easy to make, we can reduce the number of candidate entities of mentions with coarse-grained type.

%Since we observe that coarse-grained type is beneficial for Entity Linking process, we propose to interact those two tasks.
% It's confirmed that type feature is beneficial for entity linking~\cite{gupta2017entity}, it helps to reduce the candidate size and filter some entity with inconsistent type with entity mention. In general, the coarse-grained type is enough for entity linking task, it will help to raise about 3\% point on F1 score in conll 2003 dataset.
\begin{figure}[h]
\centering
\includegraphics[width=0.50\textwidth, angle=0]{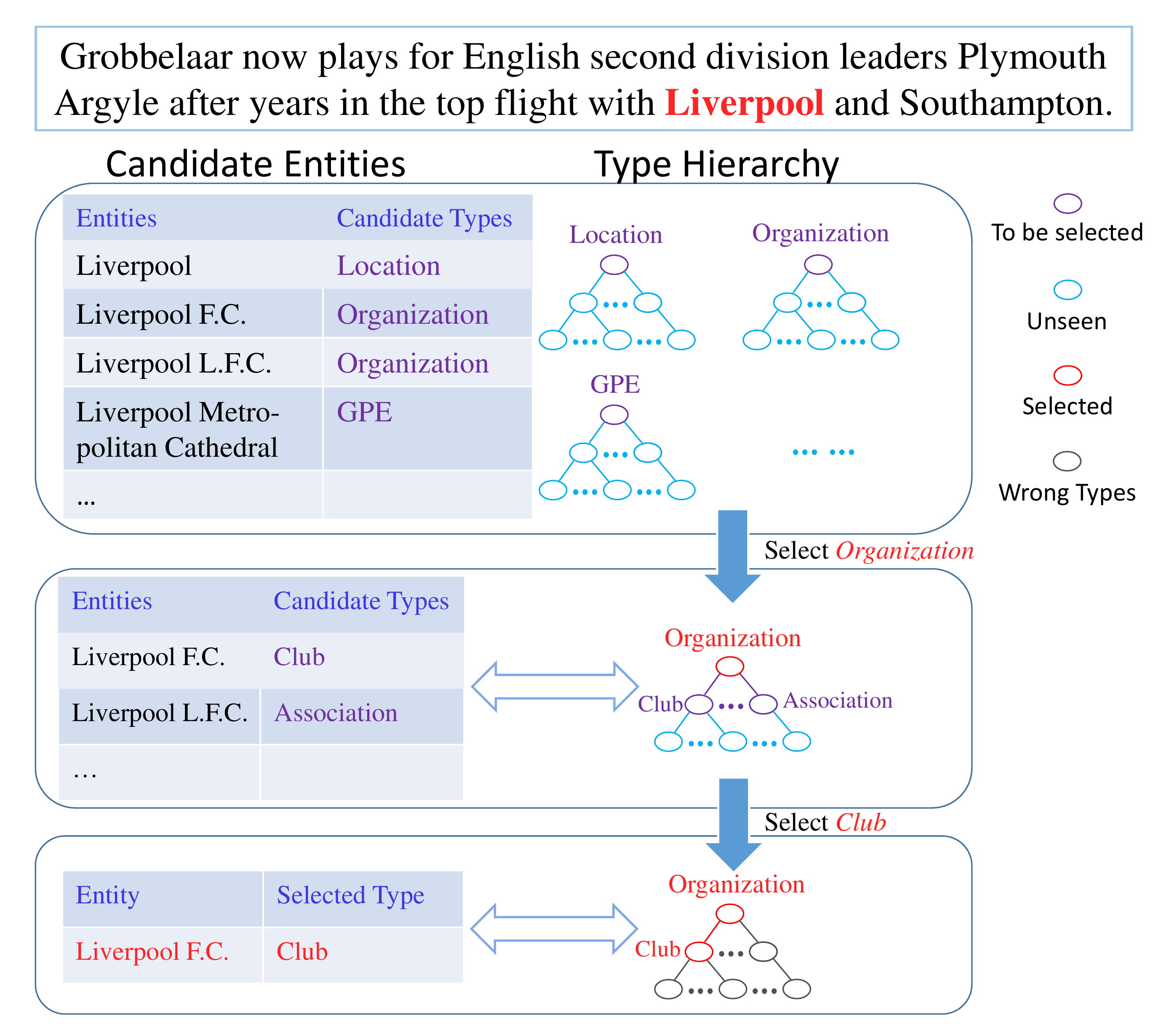}
\caption{Interaction between Typing and Linking}
\label{interaction}
\end{figure}
\begin{figure*}[t]
\centering
\includegraphics[width=0.9\textwidth, angle=0]{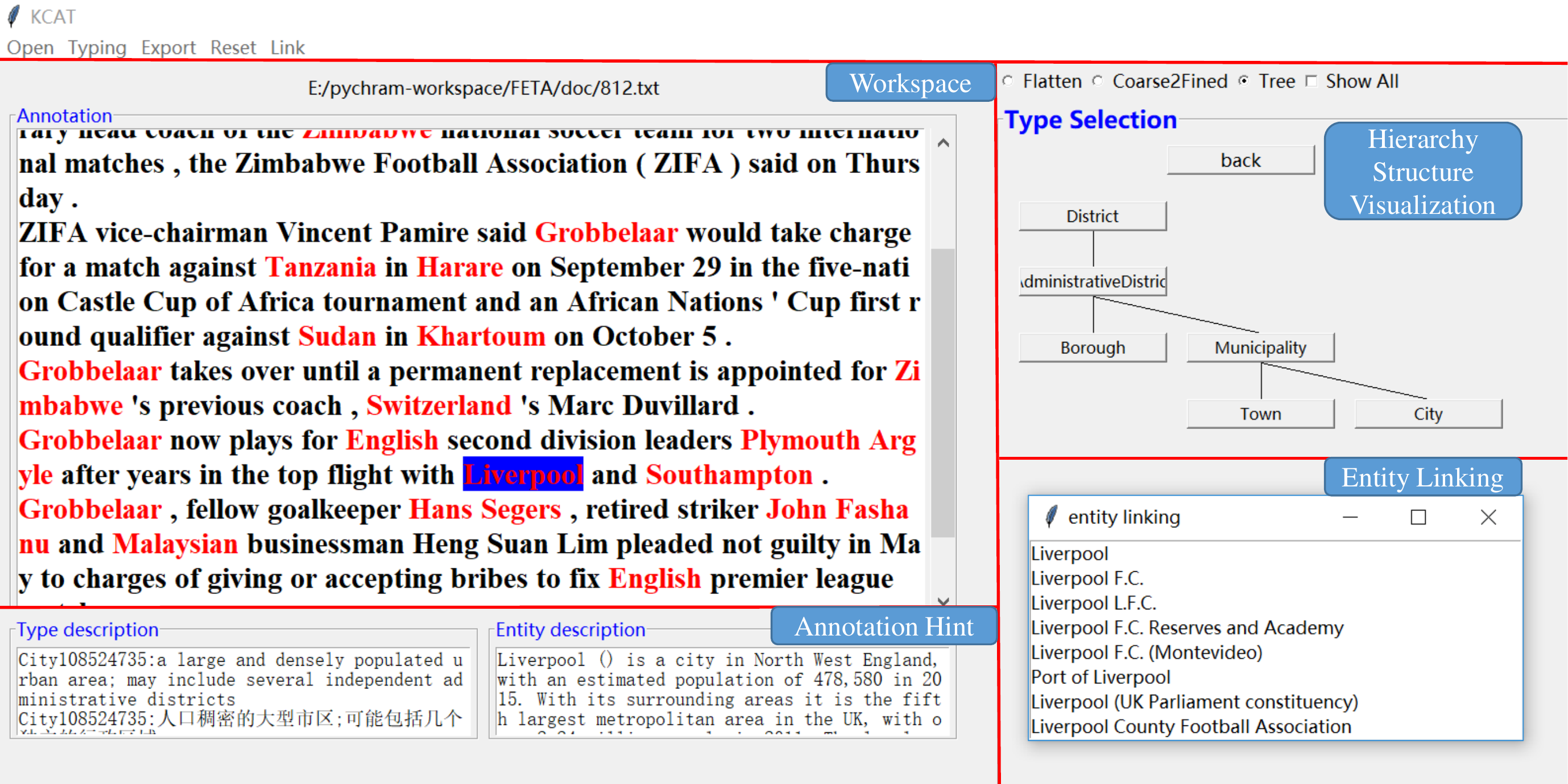}
\caption{A Screenshot of KCAT (different parts are seperated by bold red line)}
\label{screenShot}
\end{figure*}

\noindent
\subsection{Multi-step Typing}
\textbf{Entity Linking} and \textbf{Entity Typing} are mutually improved: a) EL helps to reduce the size of candidate types of ET; b) ET helps to filter out irrelevant candidate entities of EL with inconsistent types. Based on this observation, we propose a Multi-step Typing scheme. Figure \ref{interaction} demonstrates the interaction between entity typing and linking. The left part shows the candidate entities in every step and the right part shows the candidate types constrained by candidates entities. The red words``\emph{Liverpool}'' is an entity mention ``\emph{Liverpool}'', EL mistakenly links football \emph{Club} ``\emph{Liverpool F.C.}'' to ``\emph{Liverpool City}''. %EL's result, ``\emph{Liverpool City}'', contradicts the ground truth, football \emph{Club} ``\emph{Liverpool F.C.}''. It's a typical error in EL task. 
It's tough for human to label the mention as \emph{Club} without professional knowledge, but it's easy to label it as \emph{Organization}. In our scheme, the user firstly selects the coarse-grained type \emph{Organization}, and observes that the candidate entities which contain \emph{Organization}. %some football club and football association, 
Gold entity, ``Liverpool F.C.'', and \emph{Club} can be easily picked out. %, finally we select the target fine-grained type \emph{Club}. 
Through hierarchy type selection, the user focus on a few candidate entities, which can prompt user to do deeper type selection. % Through these simple interactions, the candidate size of entities and types reduces step by step, which make it easy to find the target type in $\mathcal{K}$. 
\section{KCAT}\label{KCAT ARC}
KCAT is developed based on standard Python GUI library Tkinter, hence it only needs Python installation as prerequisite. It provides user-friendly interfaces for annotators. KCAT contains two main modules: \textbf{Annotator Client} and \textbf{Manager Module}. 
\begin{itemize}[leftmargin=*]
    \item \textbf{Annotator Client:} With the help of Knowledge Constraint, KCAT makes the impossible annotation of FET possible. An Annotation Client is designed to further accelerate the annotation process. 
    % To further accelerate the annotation process, we design an Annotator Client as the front-end of KCAT.
%The Annotator Client helps annotator to accelerate the annotation process. %It provides Hierarchical Structure Visualization which helps annotator to search the target type hierarchically. 
It provides 4 practical functions to reduce annotation time: \textbf{a) Hierarchical Structure Visualization}, \textbf{b) Annotation Hint}, \textbf{c) Annotation Modification}, \textbf{d) Annotation Export}.
    \item \textbf{Manager Module: } KCAT also provides a Manager Module to analyse the annotation results.
\end{itemize}
%The main functions in Annotator Client include: a) Hierarchical Structure Visualization, b) Annotation Hint, c) Annotation Modification and d) Annotation Export. At the same time, KCAT also provides Manager Module to analyse the annotation results. 
We will introduce these two modules in following sections respectively.
\subsection{Annotator Client}
Figure \ref{screenShot} shows the interface of Annotator Client. The interface consists of 5 parts. The toolbar contains some basic functions for \textbf{Annotation Modification} and \textbf{Annotation Export}. The main area in the upper left is the text in annotating, in which mentions are colored by red and types are colored by blue. The upper right area shows the \textbf{Hierarchical Structure Visualization} and the bottom left area shows the \textbf{Annotation Hint} which helps annotators to know more about the type and entity information. The children window titled as ``entity linking'' in the bottom right shows the candidate entities ranked by entity linking system.

\noindent
\textbf{a) Hierarchical Structure Visualization}
\noindent

On most entity typing datasets, types can be formed by hierarchical structure, hence Directed Acyclic Graph (DAG) is a good view to represent this structure. By this hierarchical structure, KCAT supports up-down searching without scanning irrelevant types in other subgraph.
 
\noindent
\textbf{b) Annotation Hint}
\noindent

Annotator may not memorize all the definitions of types and descriptions of entities. Therefore, prompting type definitions and entity descriptions will contribute to the labeling process. KCAT provides the definition of type extracted from WordNet\cite{fellbaum2012wordnet} and the first paragraph of Wikipedia page related to entity to help the annotator further acquainted with the type and entity. 

\noindent
\textbf{c) Annotation Modification}
\noindent

%It is inevitable that the annotator gives incorrect annotations, 
Sometimes users need to revise the annotation when they make mistakes. %This time consumption can't be neglected, 
KCAT provides several efficient modification actions to revise these incorrect annotations.

\textbf{Action Undo and Redo:} annotators can cancel their previous actions or redo their canceled actions to return to any previous states by press the shortcut key Ctrl+z and Ctrl+y respectively.

\textbf{Label Modify and Reset:} if an entity mention receives an incorrect
type, annotator only needs to put the cursor inside the span and restart labeling. In addition, annotator can reset the annotations by shortcut key Ctrl+r.

\noindent
\textbf{d) Annotation Export}
\noindent

KCAT provides the ``Export'' function, which exports the annotated text as standard format (ended with txt) or json format so that it's easy to be processed by users.

\subsection{Manager Module}

The Manager Module aims to evaluate the quality of annotated files, analyzes the detailed disagreements of different annotators and integrates all annotation results from multiple annotators.

\noindent
\textbf{Multiple Annotators Comparison}
\noindent

\begin{figure}[h]
\centering
\includegraphics[width=0.35\textwidth,height=0.10\textwidth ]{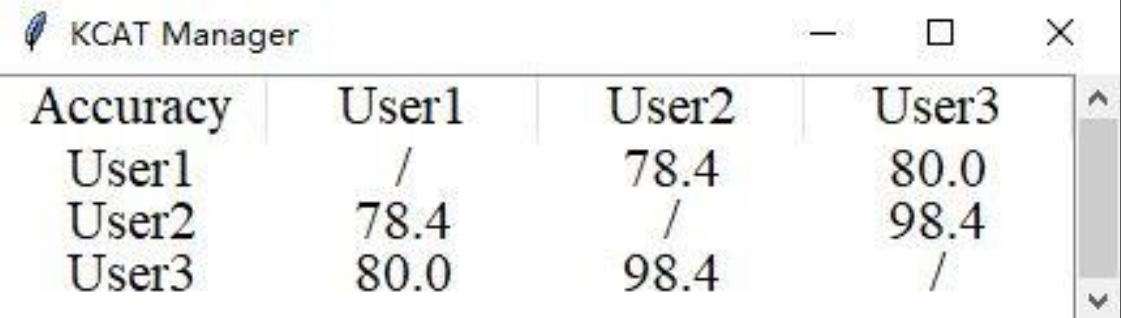}
\caption{Multiple Annotator Analysis}
\label{Multi-Annotator Analysis}
\end{figure}

The annotations from multiple annotators are inconsistent, in order to evaluate the quality of annotations, KCAT provides a Multiple Annotator Comparison interface to generate the accuracy matrix to measure consistency among multiple annotators, which is illustrated in Figure \ref{Multi-Annotator Analysis}.

\begin{figure}[h]
\centering
\includegraphics[width=0.47 \textwidth, angle=0]{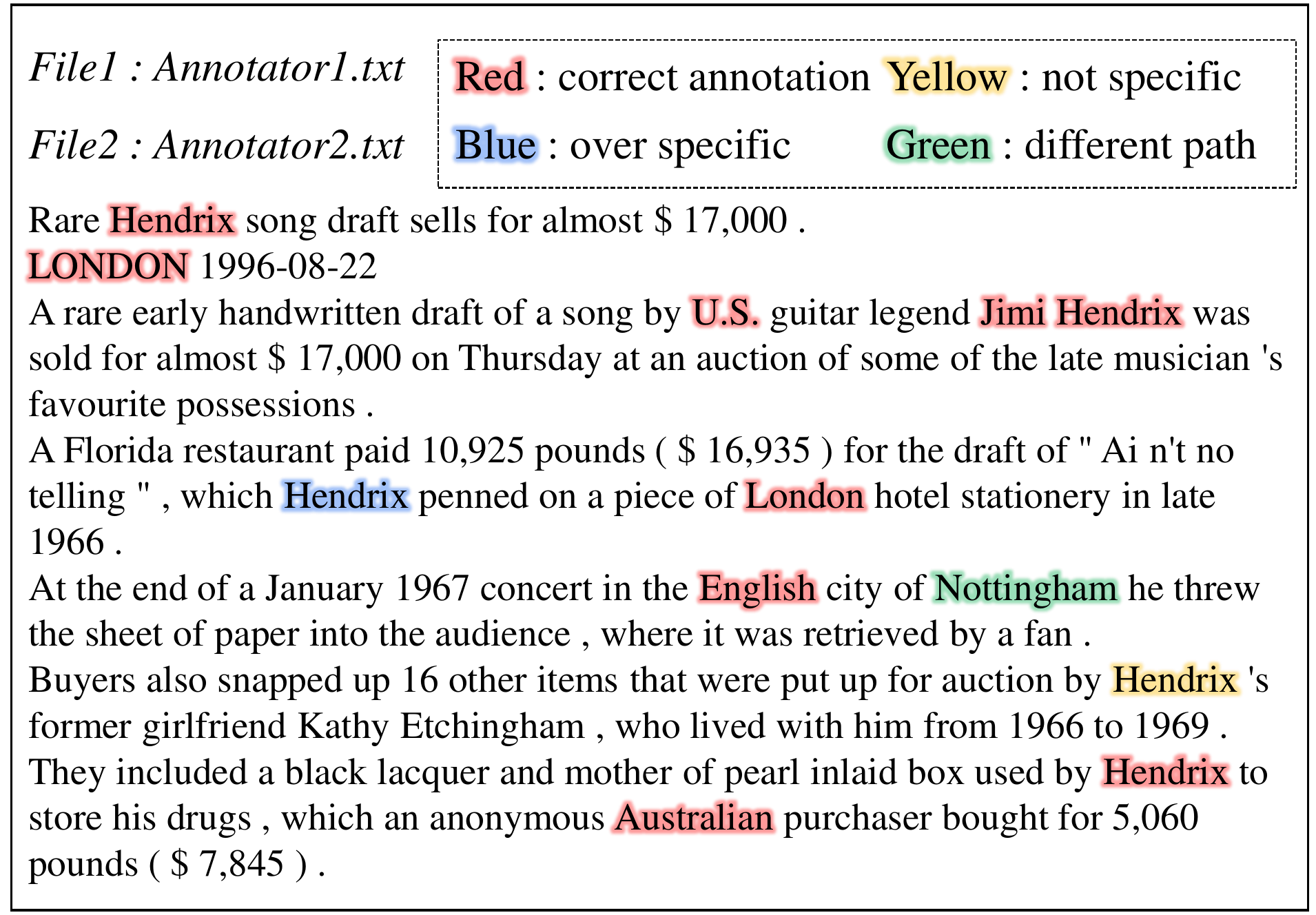}
\caption{Error Analysis}
\label{error analysis}
\end{figure}

\noindent
\textbf{Error Analysis}
\noindent

There are three common error patterns in entity typing task: (a) Over Specific;
mislabeling parent type as child type (i.e., annotated as athlete while the ground truth type is person). (b)Not Specific; in contrary to former (a). (c)Incorrect Path; labeling the wrong child type (i.e., entity is annotated as athlete while the ground truth type is artist and they are both child type of person). KCAT provides an interface to generate the error analysis report in ``.tex'' format, as shown in Figure \ref{error analysis}, different errors are rendered in different colors.

\noindent
\textbf{Annotations Integration}
\noindent

The Annotations Integration interface provides a method to integrate all annotation results from crowdsourcing annotations to generate final labels by voting.

\section{Experiment}\label{experiment}
In order to verify the efficiency of KCAT, we conduct a mock annotation experiment. 100 sentences are extracted from the English dataset of Conll 2003~\cite{sang2000introduction} as the corpus to be annotated. The entity mention spans in these sentences have been annotated. Type hierarchy is extracted from following three datasets: (1) Conll 2003; (2) BBN\cite{ren2016afet}; (3) FIGER\cite{ling2012fine}; and YAGO Knowledge Base\cite{nlp.cs.rpi.edu}. The mappings between entity and its related types are provided by these datasets. We have chosen two annotation modes: (a) without pre-linking, directly through top-down search or flatten search; (b) filtering out types that are inconsistent with entity types through entity linking.\\
\textbf{Annotation Efficiency. }In Table \ref{time consumption}, we compare the labeling time in aforementioned two modes and calculate the percentage of time saved with Entity Linking on different type set.
%We can observe that the efficiency of labeling can be significantly improved by entity linking. Comparing different datasets with different number of types,
It can be observed that with the number of types increases, time consumption increases slowly with entity linking, while without entity linking, time consumption increases exponentially. The percentage of time saved also increases as the number of types expands on different datasets. When there are thousands of types in Knowledge Base, such as YAGO Knowledge Base\cite{nlp.cs.rpi.edu}, which contains 7309 types, it's impossible for human to annotate. 
% with high quality. 

% \begin{table}[t!]
% \begin{center}r
% \small
% \begin{tabular}{|l|l|p{0.1\columnwidth}|p{0.1\columnwidth}|}
% \hline \textbf{Dataset} & \textbf{\#types} & \textbf{Time(w/o EL)} &  \textbf{Time(w/ EL)} \\ \hline
% Conll2003&	5 & 5&	6\\
% \hline
% BBN & 47 &	8&	17\\
% \hline
% FIGER &	113 & 10&	30\\
% \hline
% TAC KBP2018& 7309 &	13&	-\\
% \hline
% \end{tabular}
% \end{center}
% \caption{Statistics of Labeling time(minute)}
% \label{time consumption}
% \end{table}

\begin{figure}[h]
\centering
\includegraphics[width=0.4\textwidth]{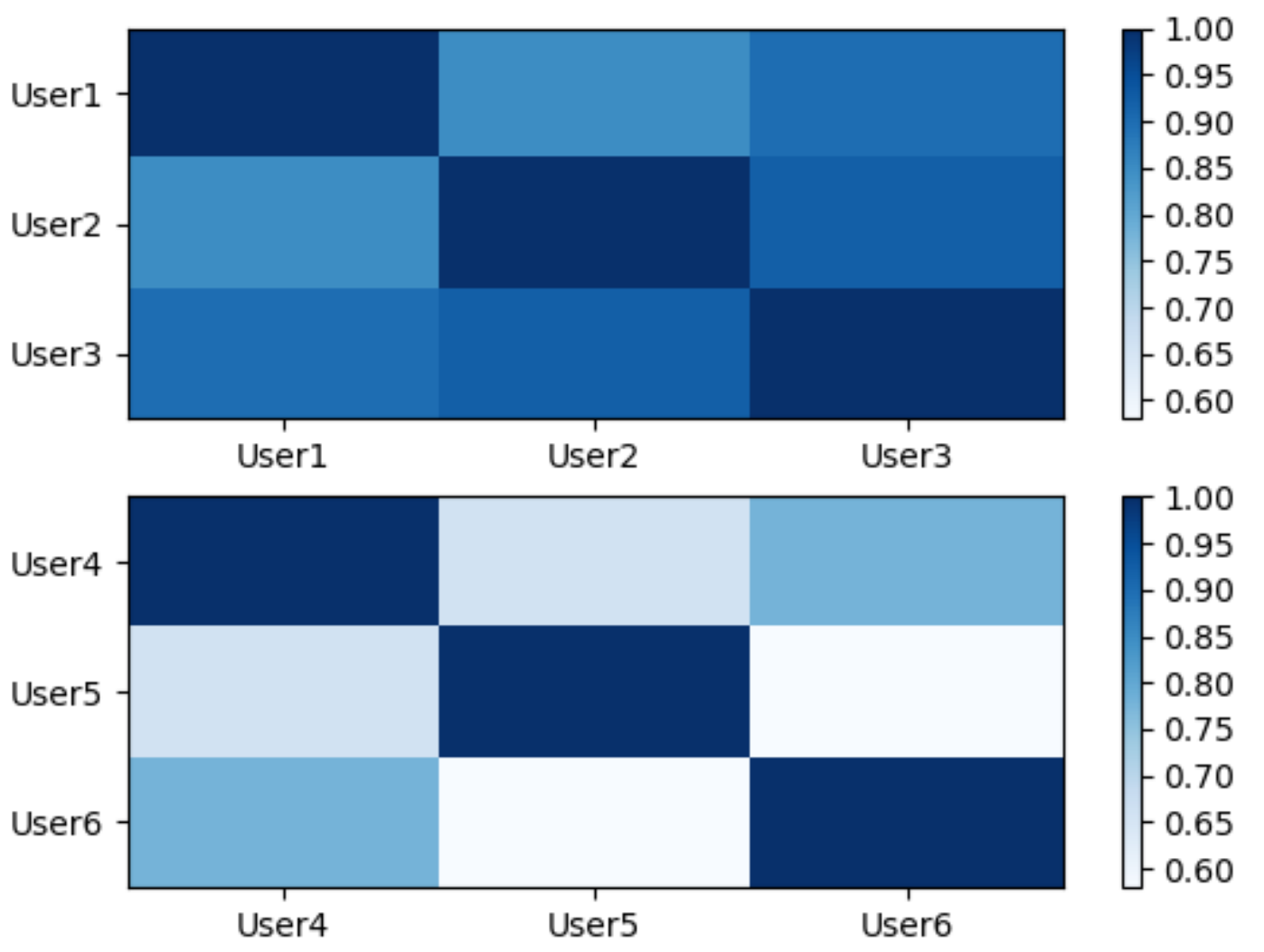}
\caption{A Comparison of Multiple Annotators in Accuracy}
\label{heat map}
\end{figure}

\begin{table}[h]
\centering
\small
\begin{tabular}{@{}lccccc@{}}
\toprule
\multirow{2}{*}{\textbf{Type Set}} & \multirow{2}{*}{\textbf{\#types}} & \multirow{2}{*}{\textbf{depth}} & \multicolumn{3}{c}{\textbf{Time}}   \\ \cmidrule(l){4-6} 
                         &            &              & \textbf{w/ EL} & \textbf{w/o EL} & \textbf{Rel\%}\\ \midrule
Conll                & 5       & 1                 & \textbf{5}                  & 6 & 6.3\%    \\
BBN                      & 47     & 2                  & \textbf{8}                  & 17 & 52.9\%   \\
FIGER                    & 112       & 2               & \textbf{10}                 & 30 & 66.7\%   \\
YAGO              & 7309                &14     & \textbf{13}                 & -  & -   \\ \bottomrule
\end{tabular}
\caption{\label{font-table} Time Consumption (minute) of Annotating 60 Sentences on Different Datasets}
\label{time consumption}
\end{table}
\noindent
\textbf{Annotation Quality. }Pairwise accuracy is used to measure the consistence between arbitrary two annotators. For multiple annotators, we can generate a heat map, each element of the heat map represents pairwise accuracy, darker color means higher accuracy. In Figure \ref{heat map}, the User 1, 2, and 3 adopt the mode (b), and the User 4, 5 and 6 adopts the mode (a), and it can be observed that the labeling quality of 1, 2, and 3 is significantly better than 4, 5, 6 as the former have higher consistency.

% Above all, KCAT helps to improve efficiency in labeling process while ensuring high quality. In addition, KCAT makes it possible to label type in difficult entity typing task with thousands of hierarchical types and the time consumption is acceptable.

% \begin{table}[]
% \small
% \centering
% \begin{tabular}{|c|c|c|c|c|}
% \hline
% Accuracy & User1 & User2 & User3 & User4\\
% \hline
% User1 & / & 72.3 & 80.0 & 58.4\\
% \hline
% User2 & 72.3 & / & 92.3 & 67.6\\
% \hline
% User3 & 80.0 & 92.3 & / & 63.0\\
% \hline
% User4 & 58.4 & 67.6 & 63.0 & /\\
% \hline
% \end{tabular}
% % \caption{Multiple Annotators Comparison in Accuracy}
% \caption{A Comparison of Multiple Annotators in Accuracy}
% \label{mutiple annotator comparison}
% \end{table}

\section{Conclusion}\label{conclusion}
In this paper, we propose an efficient Knowledge Constraint Fine-grained Entity Typing Annotation Tool, which further improves entity typing process through entity linking together with some practical functions.
%In order to reduce the workload of the annotators, we are planning to incorporate active learning into our toolkit, which helps to use annotated data more efficient.
% \section{}
% We will add active learning to our tools, 
% 主动学习 迁移到网站上,更加智能化的

\section{Acknowledgement}\label{ackknow}
This work has been supported in part by NSFC (No.61751209, U1611461), Zhejiang University-iFLYTEK Joint Research Center, Chinese Knowledge Center of Engineering Science and Technology (CKCEST), Engineering Research Center of Digital Library, Ministry of Education. Xiang Ren's research has been supported in part by National Science Foundation SMA 18-29268.

\bibliography{acl2019}
\bibliographystyle{acl_natbib}
\end{document}